\documentclass[]{article}
\usepackage{graphicx}
\usepackage{amsfonts}
\usepackage{amsmath}
\usepackage{hyperref}
\graphicspath{ {./images/} }
\usepackage{datetime}
\usepackage{caption}
\usepackage{subcaption}

\newdate{date}{13}{09}{2021}
\date{\displaydate{date}}

\title{Clustering COVID-19 Lung Scans}
\author{Jacob Householder, Andrew Householder, and John Paul Gomez-Reed\\
Faculty Mentor: Fredrick Park\\
Whittier College}

\begin{document}

\maketitle

\begin{abstract}
 With the ongoing  COVID-19 pandemic, understanding the characteristics of the virus has become an important and challenging task in the scientific community. While tests do exist for COVID-19, the goal of our research is to explore other methods of identifying infected individuals. 
 Our group applied unsupervised clustering techniques to explore a dataset of lungscans of COVID-19 infected, Viral Pneumonia infected, and healthy individuals. This is an important area to explore as COVID-19 is a novel disease that is currently being studied in detail. Our methodology explores the potential that unsupervised clustering algorithms have to reveal important hidden differences between COVID-19 and other respiratory illnesses. Our experiments use: Principal Component Analysis (PCA), K-Means++ (KM++) and the recently developed Robust Continuous Clustering algorithm (RCC). We evaluate the performance of KM++ and RCC in clustering COVID-19 lung scans using the Adjusted Mutual Information (AMI) score.
\end{abstract}

\section{Introduction}






The COVID-19 pandemic has redirected the efforts of the scientific community as a whole towards studying the characteristics of the novel virus in order to effectively limit the spread of the disease and relieve pressures on frontline medical staff. The following work is an investigation into the morphological effects of the virus. The goal is to see if one can uncover information about the similarities of symptomatic presentations of COVID-19 in afflicted individuals--in contrast to non COVID-19 Viral Pneumonia or healthy individuals via exploration of an X-ray image data set. This work focuses on using clustering as a means to identify patterns of pulmonary tissue sequelae in X-ray images which contrasts with supervised learning, where access to pre-labeled data is needed for training. In the latter case, the affliction of the patient must be known in advance. Clustering algorithms do not require such knowledge and thus have the capability to effectively group novel or even unknown conditions. Furthermore, clustering will allow greater insight into COVID-19 as it places data-points into groups based upon the similarities within their features; this will allow us to draw comparisons between COVID-19 cases and other illnesses, revealing underlying information within the data given.  

In this work, we use the new COVID-19 Radiography Database by Chowdhury et al. \cite{rahman2020covid,chowdhury2020can}. The dataset is composed of X-ray and CT scan images separated into three classes: COVID-19 cases, Viral Pneumonia cases, and Normal lungs. From here, we have applied the K-Means++ (KM++) \cite{arthur2007k} and Robust Continuous Clustering (RCC) \cite{RCC} algorithms to this database of X-ray images of patients labeled by cause to explore the structure of the dataset.

We have found that RCC indeed exhibits better clustering performance when used on the COVID-19 database versus the benchmark KM++ algorithm. It is better able to, on the whole, correctly identify which datapoints belong in the same group based upon their true labelings, showing the promise of implementing RCC in this context. 

\begin{figure}
\includegraphics[width=\textwidth]{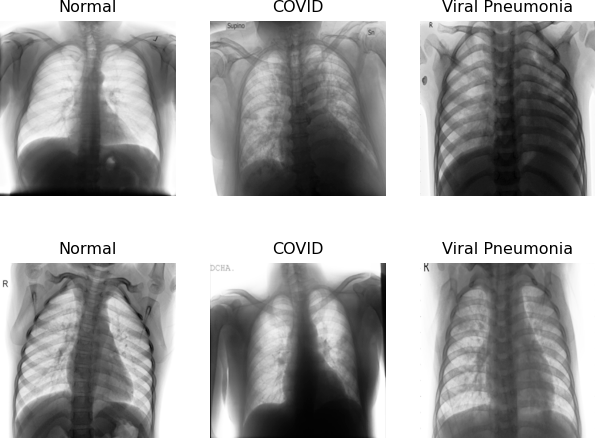}
\caption{}
\centering
\label{fig:covidDemo}
\end{figure}

Visualizing a high dimensional dataset is a difficult challenge, but doing so can help to gain intuition about any potential structure of the dataset. Recently t-Distributed Stochastic Neighbor Embedding (t-SNE) was introduced by van der Maaten and Hinton\cite{TSNE}. This method maps high dimensional data to a lower dimensional representation while attempting to preserve any pairwise neighbor structure present in the dataset. Principal Component Analysis (PCA) \cite{pearson1901lines,hotelling1933analysis} is often used for preliminary dimensionality reduction. The combination of PCA and t-SNE allows us to view the similarity of the lung scans clusters in 2 and 3 dimensions. More details and related work to t-SNE can be found in \cite{hinton2003stochastic,van2009learning,van2012visualizing,maaten2014accelerating}.

The Adjusted Mutual Information (AMI) \cite{meilua2007comparing,vinh2009information,vinh2010information} score is used to evaluate the clustering performance of RCC and KM++ in our experiments. AMI is useful as a comparative measurement for the following experiments due to its adjustment for agreement by chance.

\section{Methods}
\label{sec:methods}

We make use of four main techniques in our experiments: Principal Component Analysis (PCA) for dimensionality reduction, t-Distributed Stochastic Neighbor Embedding (t-SNE) for visualization, and K-Means++ (KM++) and Robust Continuous Clustering (RCC) for clustering.

\subsection{Principal Component Analysis (PCA)}
Principal component analysis is a data analysis technique which looks to extract information from a dataset via a change of basis along the orthogonal principal components of the data. Relevantly, PCA highlights the directions of the greatest variance within the data along the main principal components. Thus, allowing for a great amount of information to be captured in a reduced number of dimensions. Intuitively this is done by calculating the direction of the greatest variance within the data which is the first principal component. Then, the remaining principal components are found along the direction of the vector with the next greatest variance with the condition that it is orthogonal to the other principal components. Since the most variability will be captured by the first principal components, PCA is often used to reduce dimensionality by neglecting the later principal components which should hold much less information. This makes PCA a relevant technique for processing large, high-dimensionality datasets while maintaining fidelity of information \cite{jolliffe2016pca}.

\subsection{K-Means++ (KM++)}
To serve as a benchmark comparison for our experimentation we used the the well-known unsupervised clustering algorithm K-Means++ (KM++). 
The KM++ algorithm improves upon the clustering performance of the standard K-Means algorithm while preserving its speed and simplicity \cite{arthur2007k}. 
The standard K-Means Algorithm originally proposed by Lloyd \cite{lloyd} looks to cluster a given dataset $\chi = \{x_1, x_2, ..., x_n\}$ by selecting $k$ initial cluster centers $C = \{c_1, c_2, ..., c_k\}$ as arbitrary datapoints from a set of data for a predetermined $k$. 
The iterative method then proceeds by assigning each datapoint $x_i\in\chi$ to the nearest cluster center.
The position of the cluster center is then adjusted to be the center of mass of each point in cluster $C_i$: $$c_i = \frac{1}{|C_i|}\sum_{x\in{C_i}}x.$$ 
This process of assigning points to the nearest center and recomputing center location continues until the composition of the clusters (and thereby the center of mass of each cluster) no longer changes. 

Now, KM++ sets out to account for the fact that, on its own the K-Means algorithm is initialization sensitive--performance of the algorithm in general depends upon the random initial selection of the first cluster centers. 
This can lead to suboptimal clustering performance, as shown by Arthur and Vassilvitskii \cite{arthur2007k}. 
Instead, it is proposed that initial cluster centers are chosen in a more calculated manner, by first picking one point at random be the first of $k$ centers, then picking the next center to be a another datapoint with the probability: $$\frac{D(x)^2}{\sum_{x\in\chi}D(x)^2}$$ given that $D(x)$ is the straight-line distance between the closest center already chosen and $x\in\chi$ is an element of our dataset. 
This process is repeated until all initial $k$ cluster centers are chosen at which point the standard K-Means interative process is employed. 
Choosing cluster centers in this manner encourages greater distances between the initial cluster centers, giving the algorithm a better chance of identifying separable clusters. 
 With its more informed method of intialization, Arthur and Vassilvitskii show that KM++ indeed does tend to outperform K-Means \cite{arthur2007k}.

\subsection{Robust Continuous Clustering (RCC)}

RCC is a recently developed clustering technique that evolves a continuous representation of the input data set such that similar data points form tight clusters \cite{RCC}. The objective function is as follows:

$$C(U) = \frac{1}{2}\sum^{n}_{i=1}{||x_i - u_j||_2^2} + \frac{\lambda}{2}\sum_{p,q\in\xi}{w_{p,q}\rho(||u_p - u_q||_2)}$$

$$\rho(y) = \frac{\mu{y^2}}{\mu + y^2}$$
where $X = [x_1,x_2,\dots,x_i,\dots]$, $x_i \in \mathbb{R}^D$, and $U$ is the set of representative points, which are initially set to $X$. Note, $\xi$ is a set of edges connecting data points to one another. This edge set is constructed via mutual-K Nearest Neighbors (m-KNN). Mutual-K Nearest Neighbors adds the extra criteria that two points are only neighbors if they are in each others set of $k$ nearest neighbors. An interesting feature of RCC is that the number of clusters is not an explicit parameter. This contrasts with KM++, wherein this value is fixed for the duration of the algorithm. $\rho$ is used to represent an estimator function, in our case we use the Geman-McClure estimator. Below is another form of the RCC objective function, which takes into account the connections formed by m-KNN:
$$C(U,\mathbb{L}) = \frac{1}{2}\sum^{n}_{i=1}{||x_i - u_j||_2^2} + \frac{\lambda}{2}\sum_{p,q\in\xi}{w_{p,q}(l_{p,q}||u_p - u_q||_2 + \Psi(l_{p,q}))}$$

$$\Psi(l_{p,q})=\mu(\sqrt{l_{p,q}}-1)^2.$$
RCC can be optimized via alternating minimization, thus:

$$\underset{U}{\arg\min} \frac{1}{2}||X-U||_F^2 + \frac{\lambda}{2}\sum_{(p,q)\in \xi}{w_{p,q}l_{p,q}(e_p-e_q)(e_p-e_q)^\top}$$
the optimal value of $l_{p,q}$ becomes:
$$l_{p,q} = \left( \frac{\mu}{\mu ||u_p - u_q||_2^2}\right)^2.$$

\section{Metric - Adjusted Mutual Information}
Mutual Information (MI) is an entropy based measure which quantifies the amount of information given by a random variable in a particular clustering based on the probability of a particular point lying in any given cluster. The formulation of MI is given by:
$$MI(U,V)=\sum_{i=1}^{R}\sum_{i=1}^{C}{P_{UV}(i,j)log\frac{P_{UV}(i,j)}{P_{U}(i)P_{V}(j)}}$$
where $U_{i}$ and $V_{j}$ are clusters in separate partitions of the same set of data ranging from $\{U_{1},...,U_{R}\}$ and $\{V_{1},...,V_{C}\}$ respectively. The probability of a data point lying in a given cluster is denoted by $P_{U}$ and $P_{V}$, and the joint probability between partitions is labeled $P_{UV}$. 

To obtain the Adjusted Mutual Information (AMI) score from the MI score, the adjustment of an index for chance proposed by Hubert and Arabie \cite{Adjusted} is applied to the formulation of the MI:
$$\mbox{Adjusted}\_\mbox{index}=\frac{\mbox{index}\, - \, \mbox{expected}\_\mbox{index}}{\mbox{max}\_\mbox{index} \, - \, \mbox{expected}\_\mbox{index}}.$$

An AMI score of zero would indicate that clustering performance matches the prediction of simply assigning labels by chance--and thus AMI may be negative. An AMI score of one indicates a perfect labeling. Performance was measured using AMI as a metric throughout this paper. AMI was calculated using the ground truth labeling of the dataset and the labeling produced by each algorithm to evaluate clustering performance.
This method is useful for measuring the performance of clustering methods as these techniques are exploratory and do not produce necessarily meaningful class labelings in the same way supervised classification does.

\section{Results}

For each experiment the dataset was resized to $128 \times 128$ pixels, then min-max scaled, and then PCA was applied to further reduce the dimensionality. The PCA dimension was chosen such that it preserved $95\%$ of the variance of the dataset for each dataset. 
This data was then processed separately with KM++ and RCC. A search was then performed over a range of values of $k$ for KM++ and $k$ for the m-KNN routine of RCC.


\begin{figure}[h!]
\centering
\begin{subfigure}[b]{.85\textwidth}
    \includegraphics[width=\textwidth]{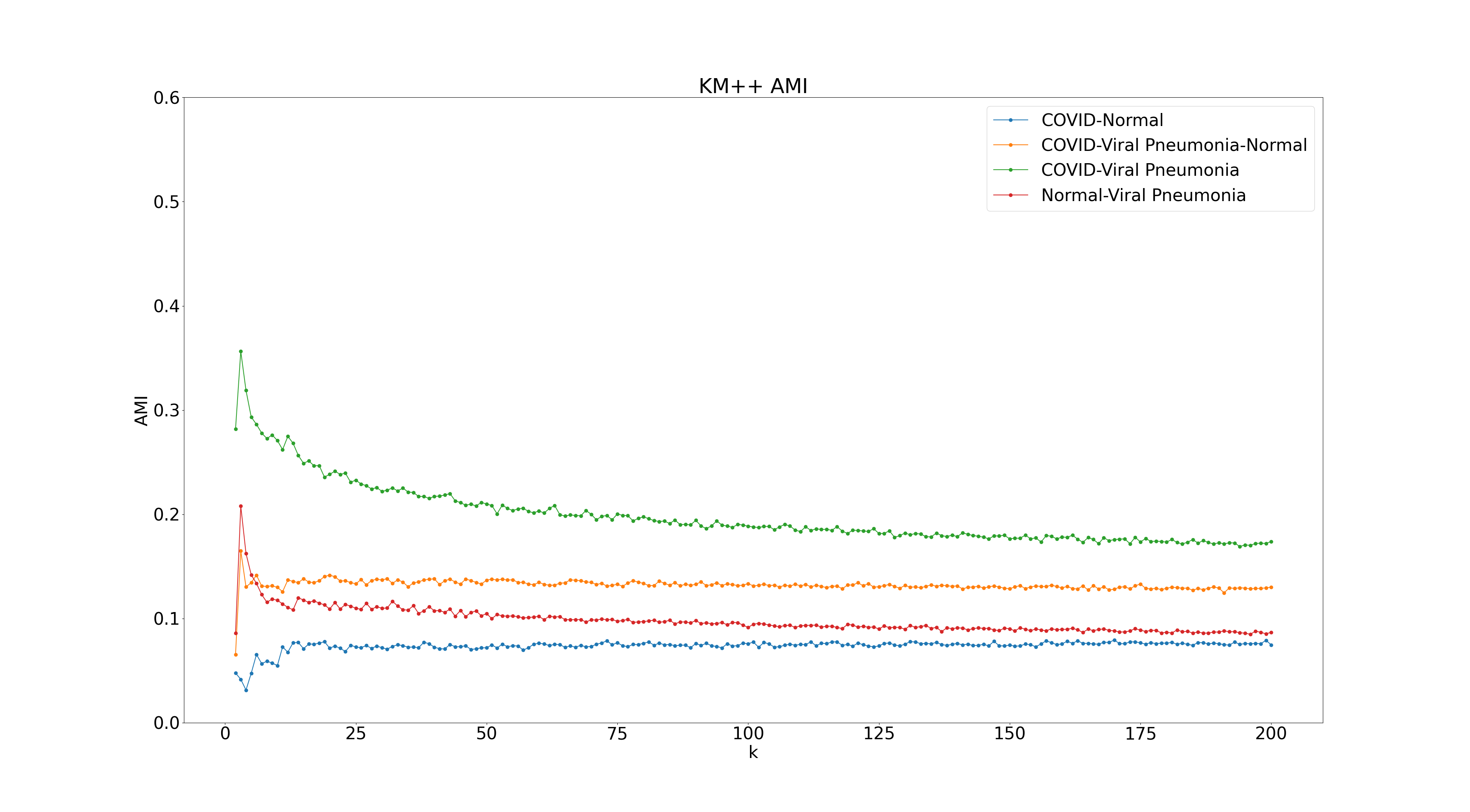}
    \caption{AMI vs. $k$, the number of clusters in KM++}
    \centering
    \label{fig:kmeans_solo}
\end{subfigure}
\begin{subfigure}[b]{.85\textwidth}
    \includegraphics[width=\textwidth]{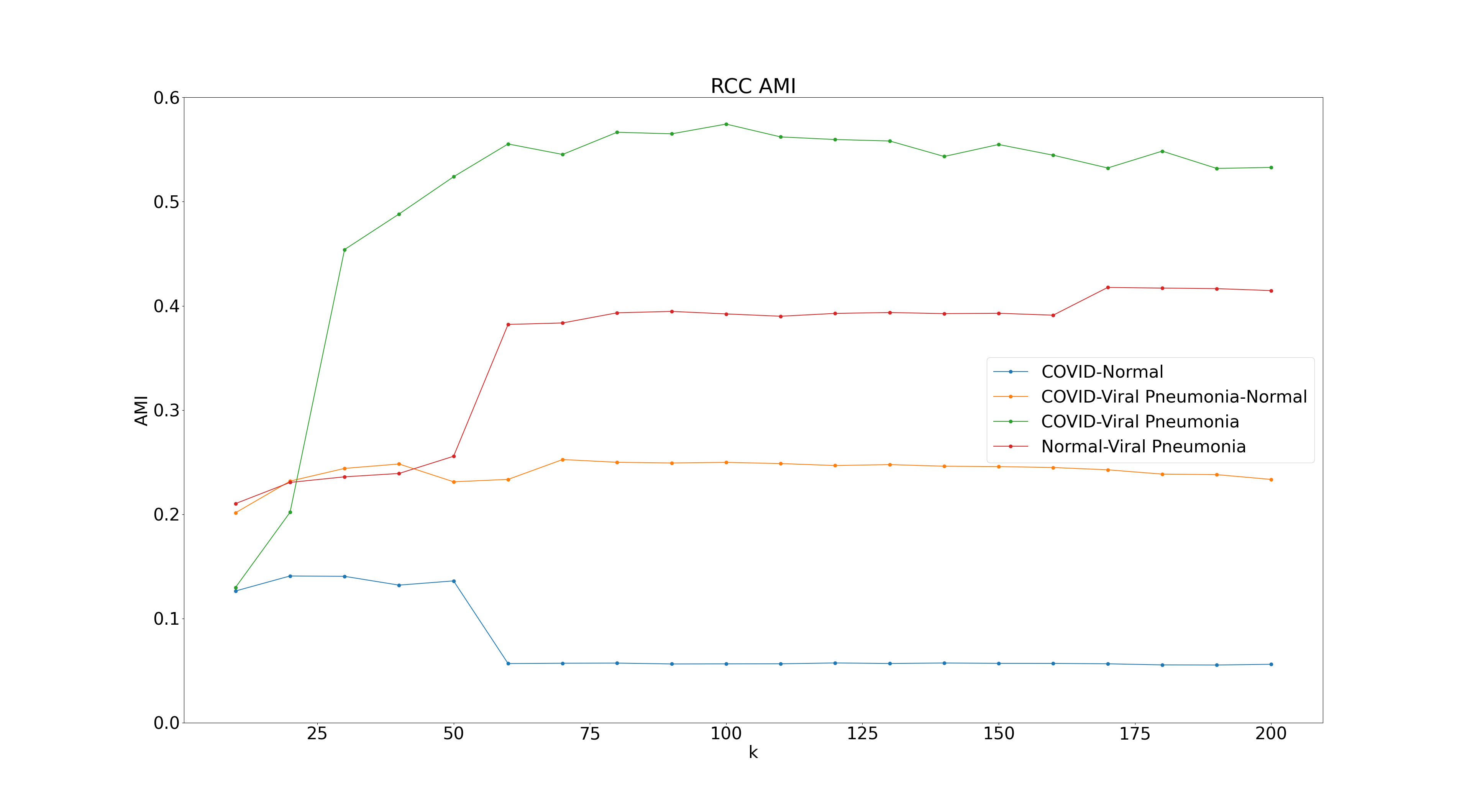}
    \caption{AMI vs. $k$, the number of nearest neighbors in RCC's m-KNN graph}
    \centering
    \label{fig:rcc_solo}
\end{subfigure}
\caption{}
\label{fig:ami_plots}
\end{figure}

\begin{table}[h!]
\centering
\begin{tabular}{||c c c||} 
 \hline
 Classes & $k$ & AMI \\ [0.5ex] 
 \hline\hline
 Covid-Normal & 170 & 0.079357 \\
 Covid-Viral Pneumonia & 3 & 0.356376 \\
 Normal-Viral Pneumonia & 3 & 0.207916 \\
 Covid-Viral Pneumonia-Normal & 2 & 0.164900 \\ [1ex] 
 \hline
\end{tabular}
\caption{Best results from the KM++ experiments.}
\label{table:1}
\end{table}

\begin{table}[h!]
\centering
\begin{tabular}{||c c c||} 
 \hline
 Classes & $k$ & AMI \\ [0.5ex] 
 \hline\hline
 Covid-Normal & 20 & 0.140792 \\
 Covid-Viral Pneumonia & 100 & 0.574444 \\
 Normal-Viral Pneumonia & 170 & 0.417678 \\
 Covid-Viral Pneumonia-Normal & 70 & 0.252518 \\ [1ex] 
 \hline
\end{tabular}
\caption{Best results from the RCC experiments.}
\label{table:2}
\end{table}

\begin{figure}[h!]
\centering
\begin{subfigure}[b]{.333\textwidth}
  \centering
  \includegraphics[width=\textwidth]{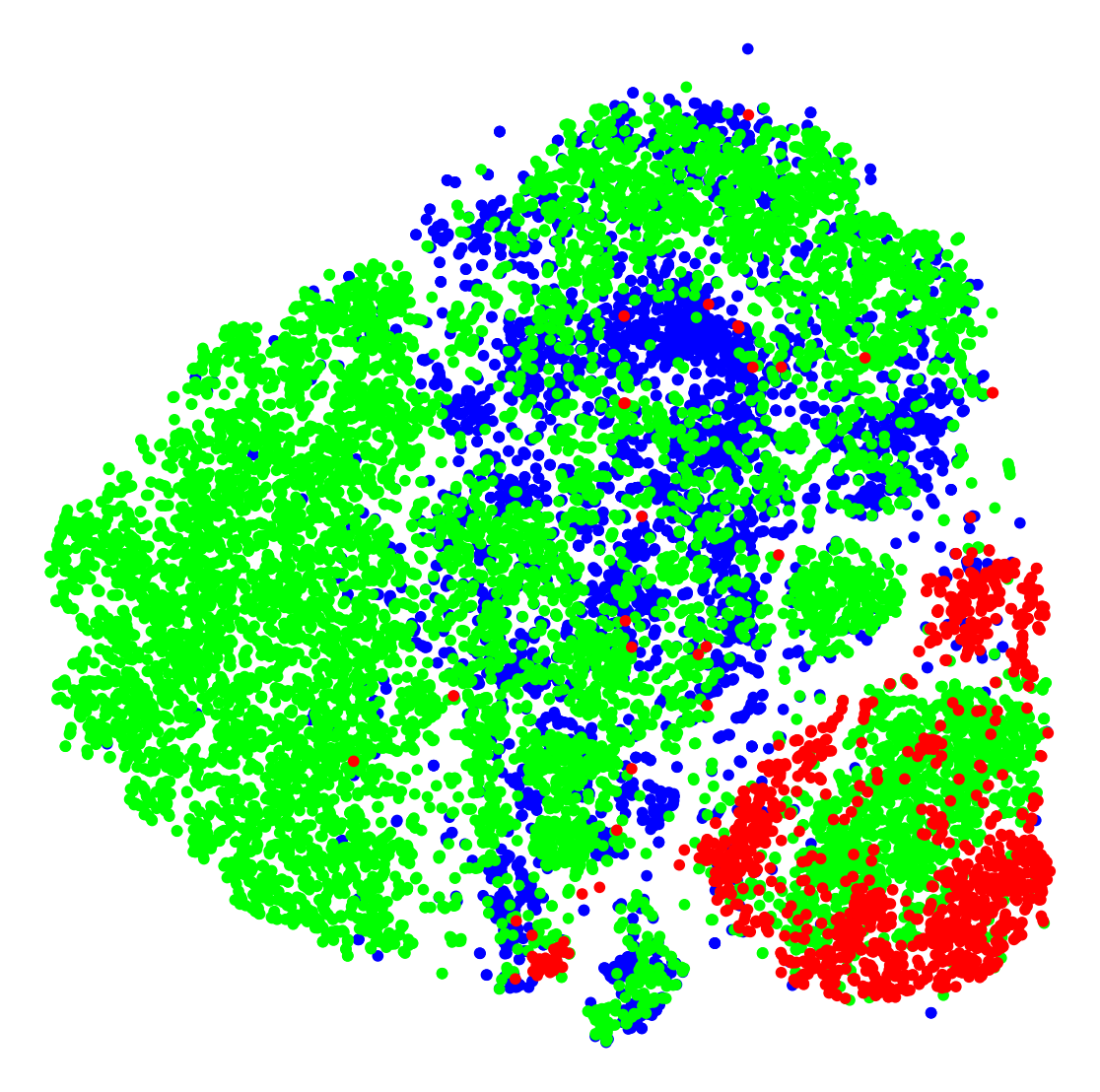}
  \caption{Ground truth}
  \label{fig:sub1}
\end{subfigure}%
\hfill
\begin{subfigure}[b]{.333\textwidth}
  \centering
  \includegraphics[width=\textwidth]{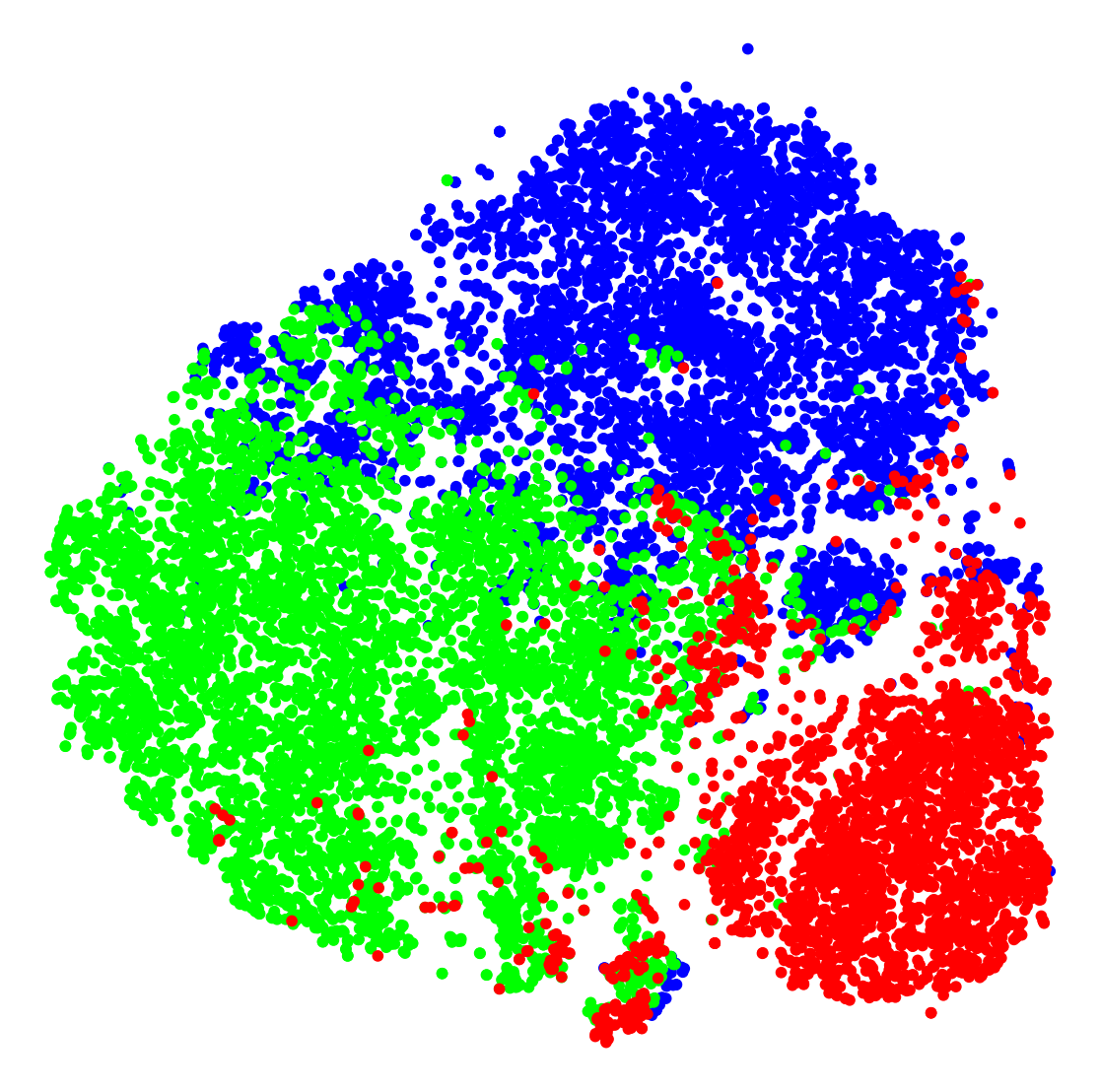}
  \caption{KM++}
  \label{fig:sub2}
\end{subfigure}%
\hfill
\begin{subfigure}[b]{.333\textwidth}
  \centering
  \includegraphics[width=\textwidth]{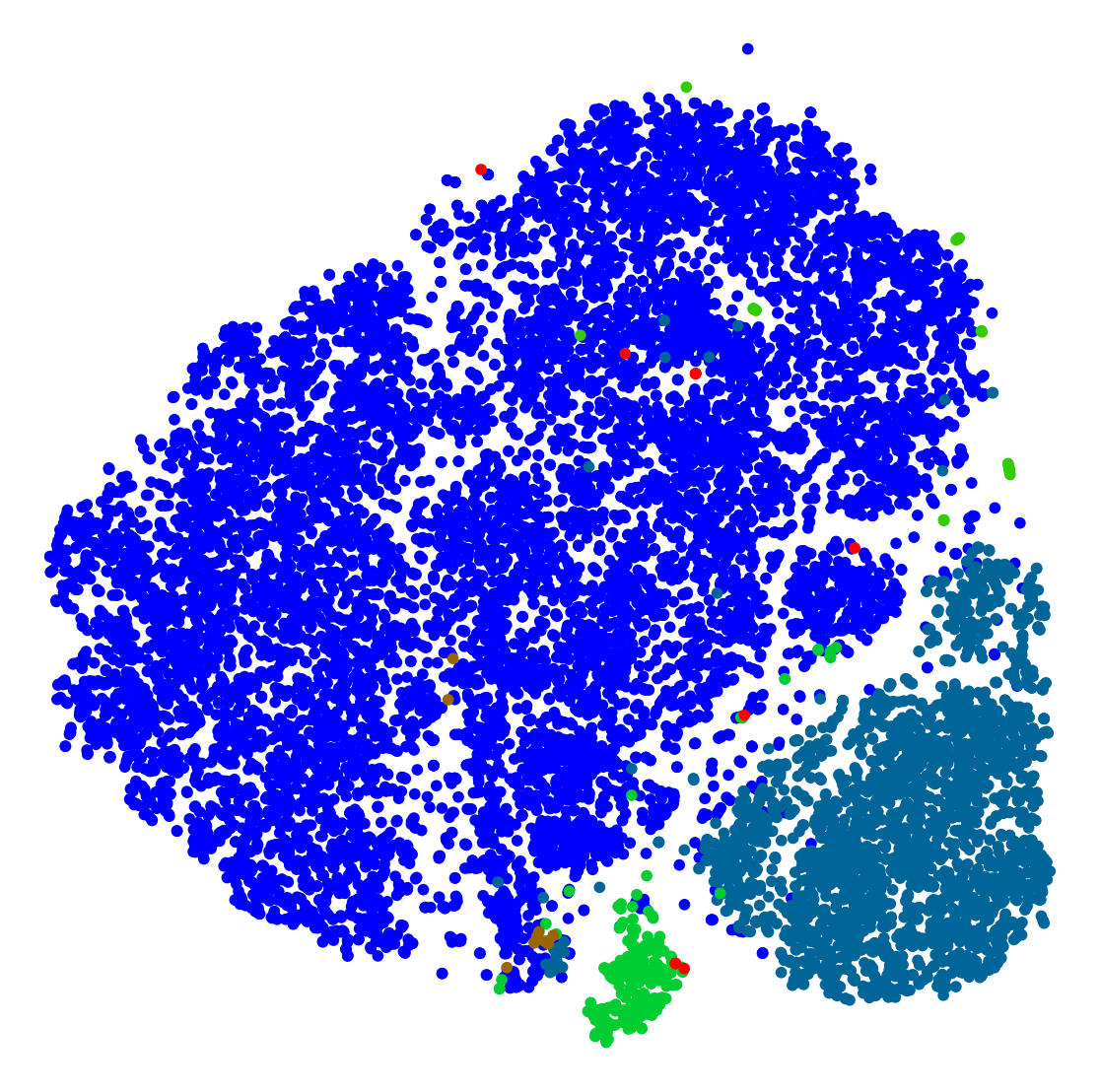}
  \caption{RCC}
  \label{fig:sub2}
\end{subfigure}%
\caption{COVID, Viral Pneumonia, and Normal t-SNE visualization}
\label{fig:3class_tsne}
\end{figure}

\begin{figure}[h!]
\centering
\begin{subfigure}[b]{.333\textwidth}
  \centering
  \includegraphics[width=\textwidth]{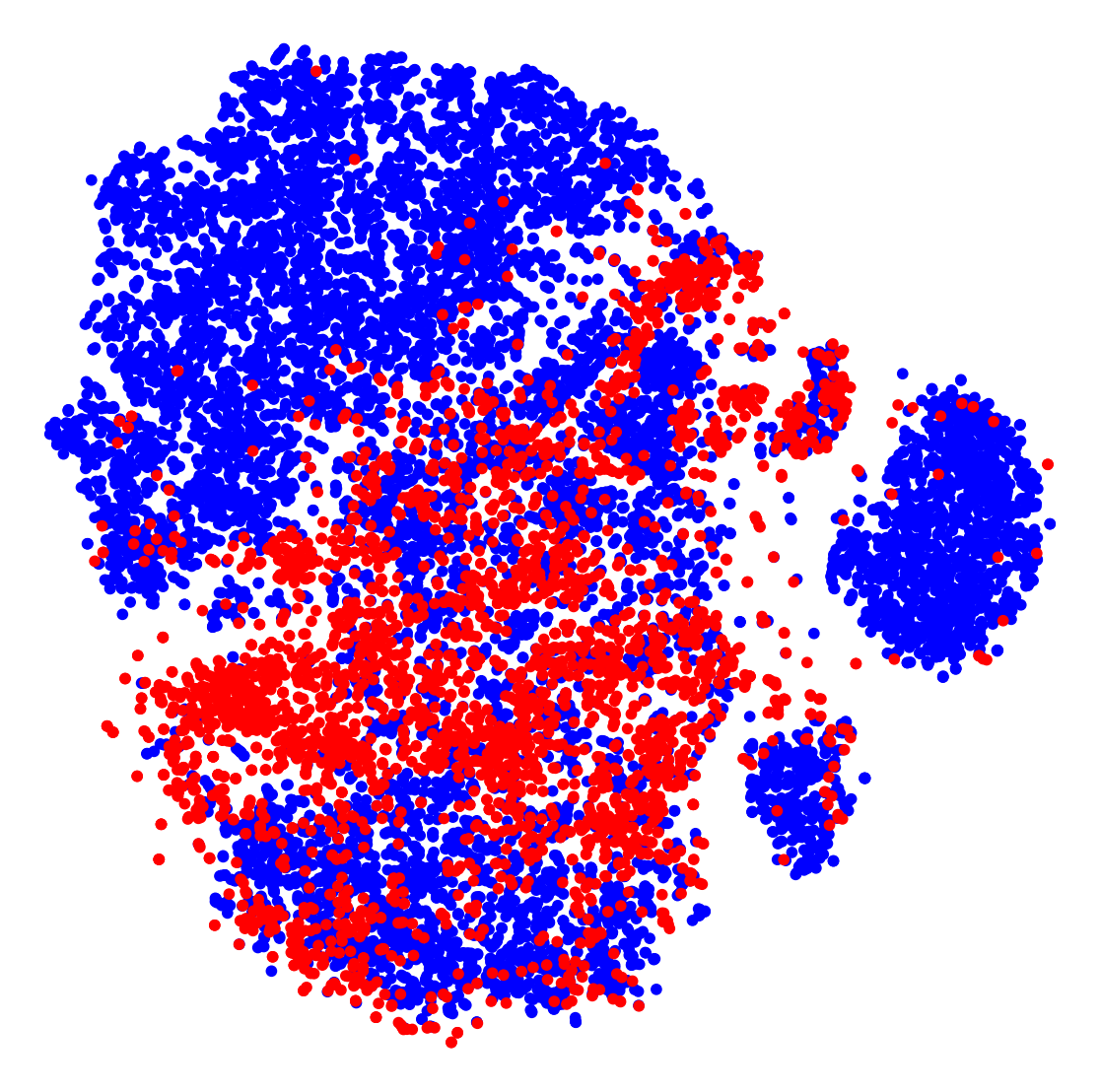}
  \caption{Ground truth}
  \label{fig:COVID-Normal_tsne(a)}
\end{subfigure}%
\hfill
\begin{subfigure}[b]{.333\textwidth}
  \centering
  \includegraphics[width=\textwidth]{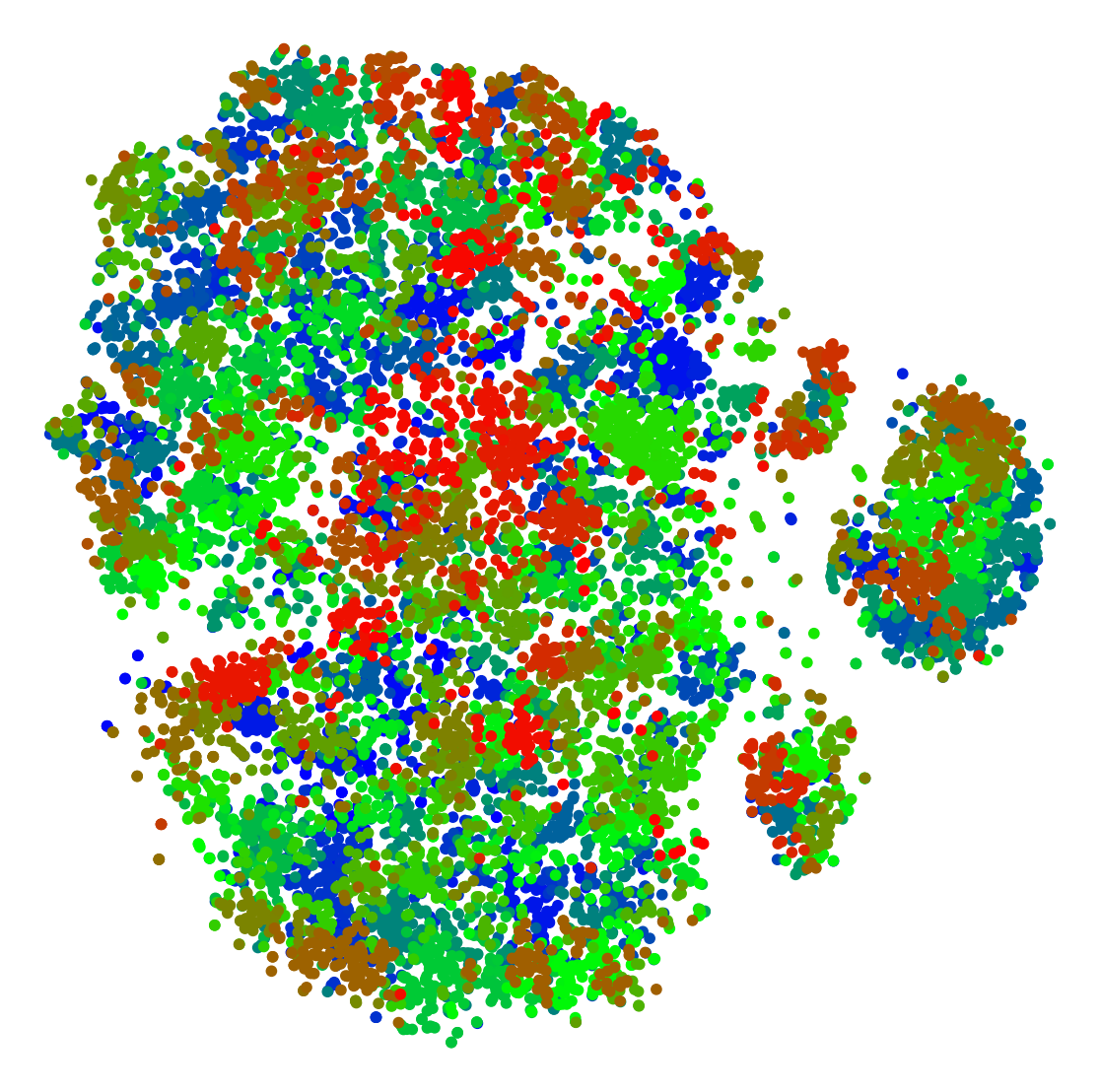}
  \caption{KM++}
  \label{fig:sub2}
\end{subfigure}%
\hfill
\begin{subfigure}[b]{.333\textwidth}
  \centering
  \includegraphics[width=\textwidth]{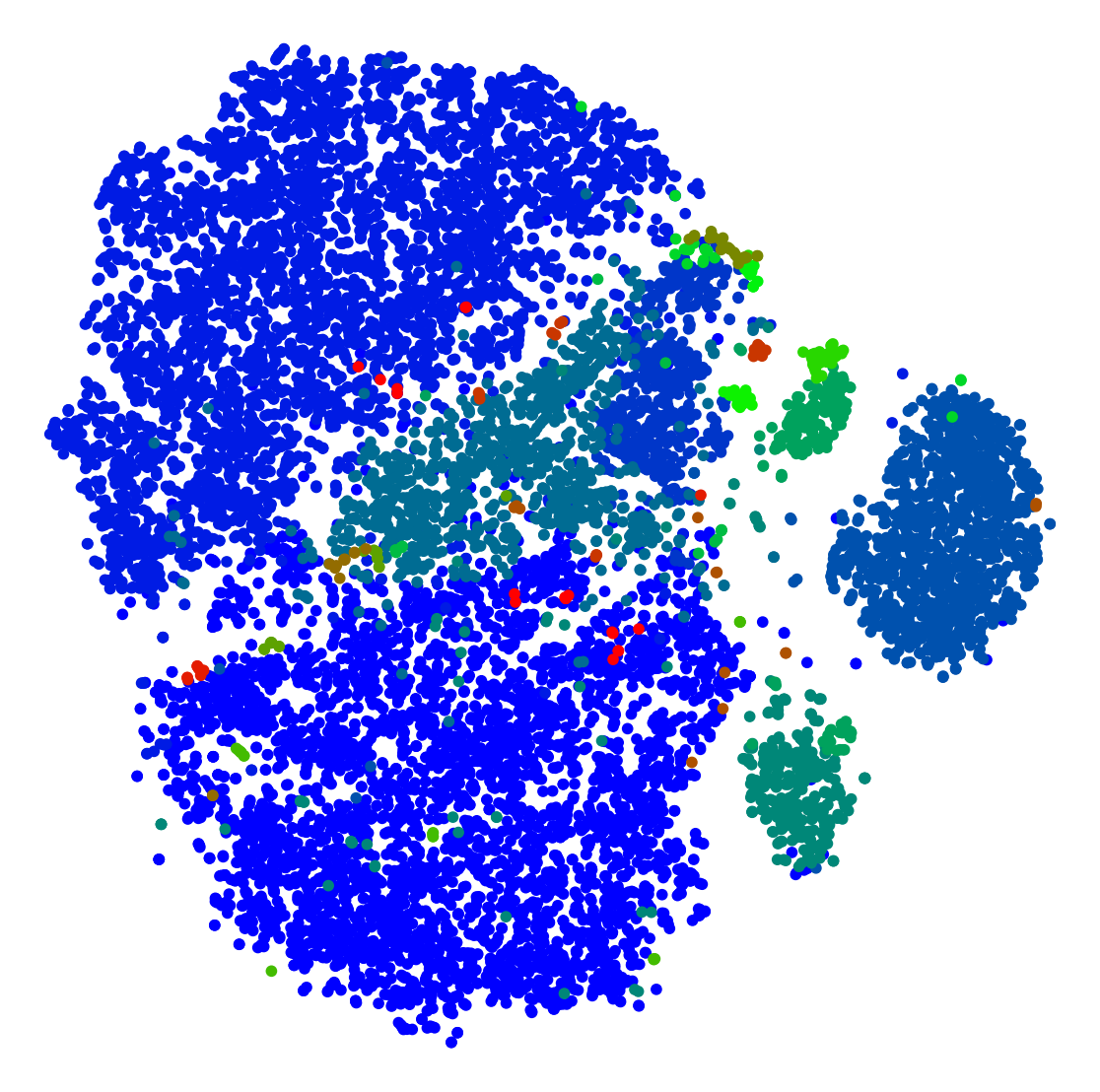}
  \caption{RCC}
  \label{fig:sub2}
\end{subfigure}%
\caption{COVID vs. Normal t-SNE visualization}
\label{fig:COVID-Normal_tsne}
\end{figure}

\begin{figure}[h!]
\centering
\begin{subfigure}[b]{.333\textwidth}
  \centering
  \includegraphics[width=\textwidth]{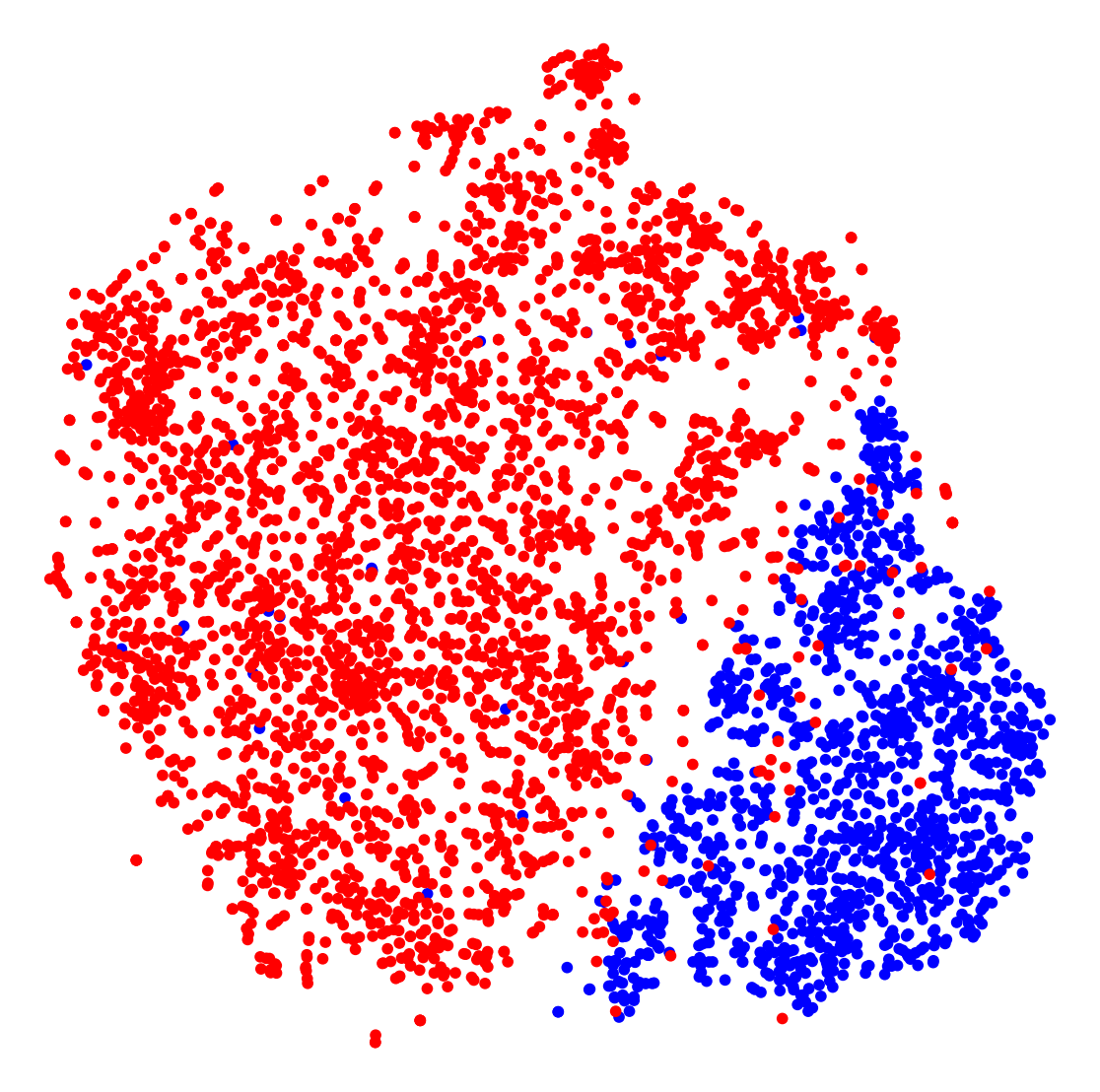}
  \caption{Ground truth}
  \label{fig:sub1}
\end{subfigure}%
\hfill
\begin{subfigure}[b]{.333\textwidth}
  \centering
  \includegraphics[width=\textwidth]{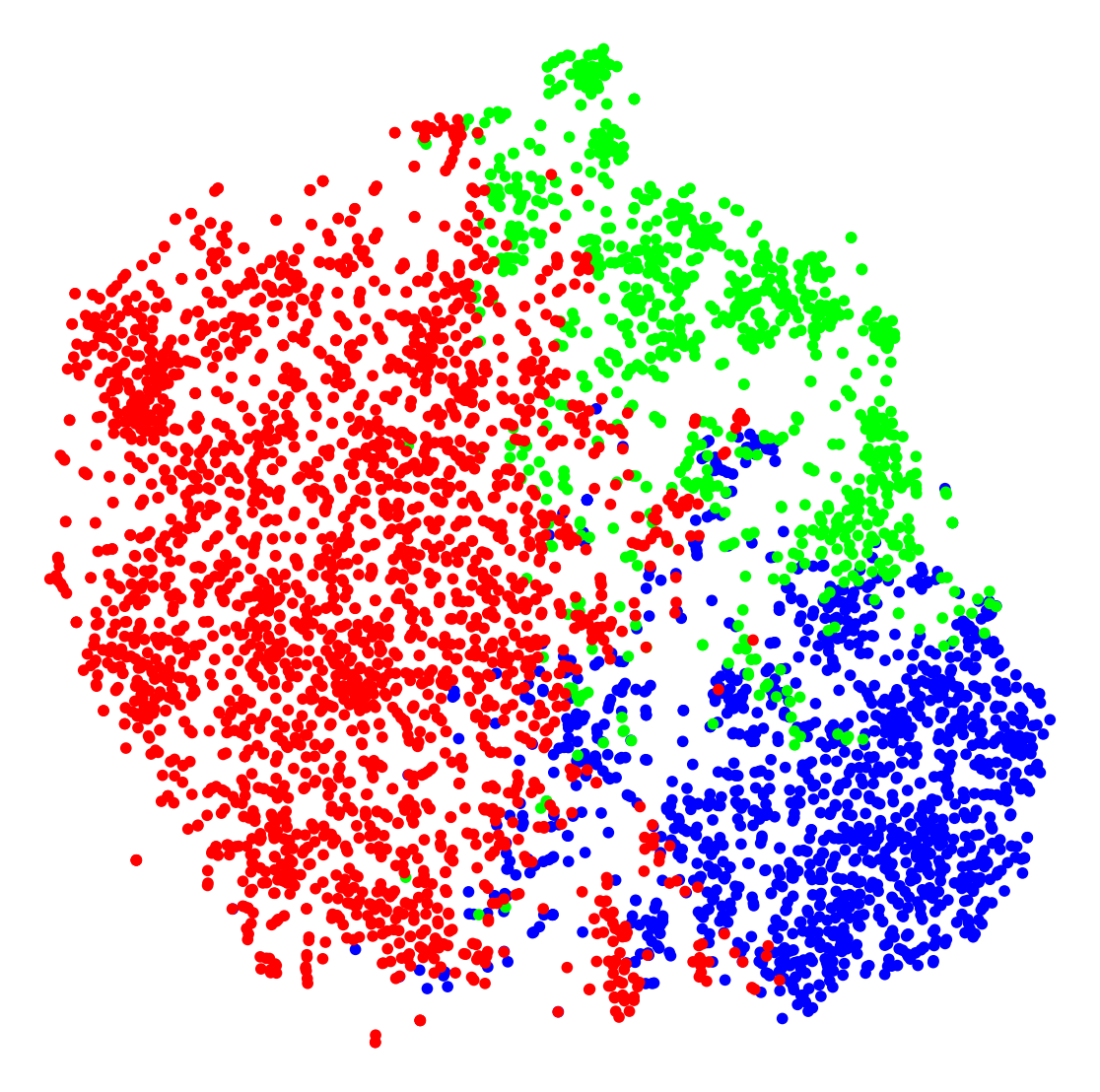}
  \caption{KM++}
  \label{fig:sub2}
\end{subfigure}%
\hfill
\begin{subfigure}[b]{.333\textwidth}
  \centering
  \includegraphics[width=\textwidth]{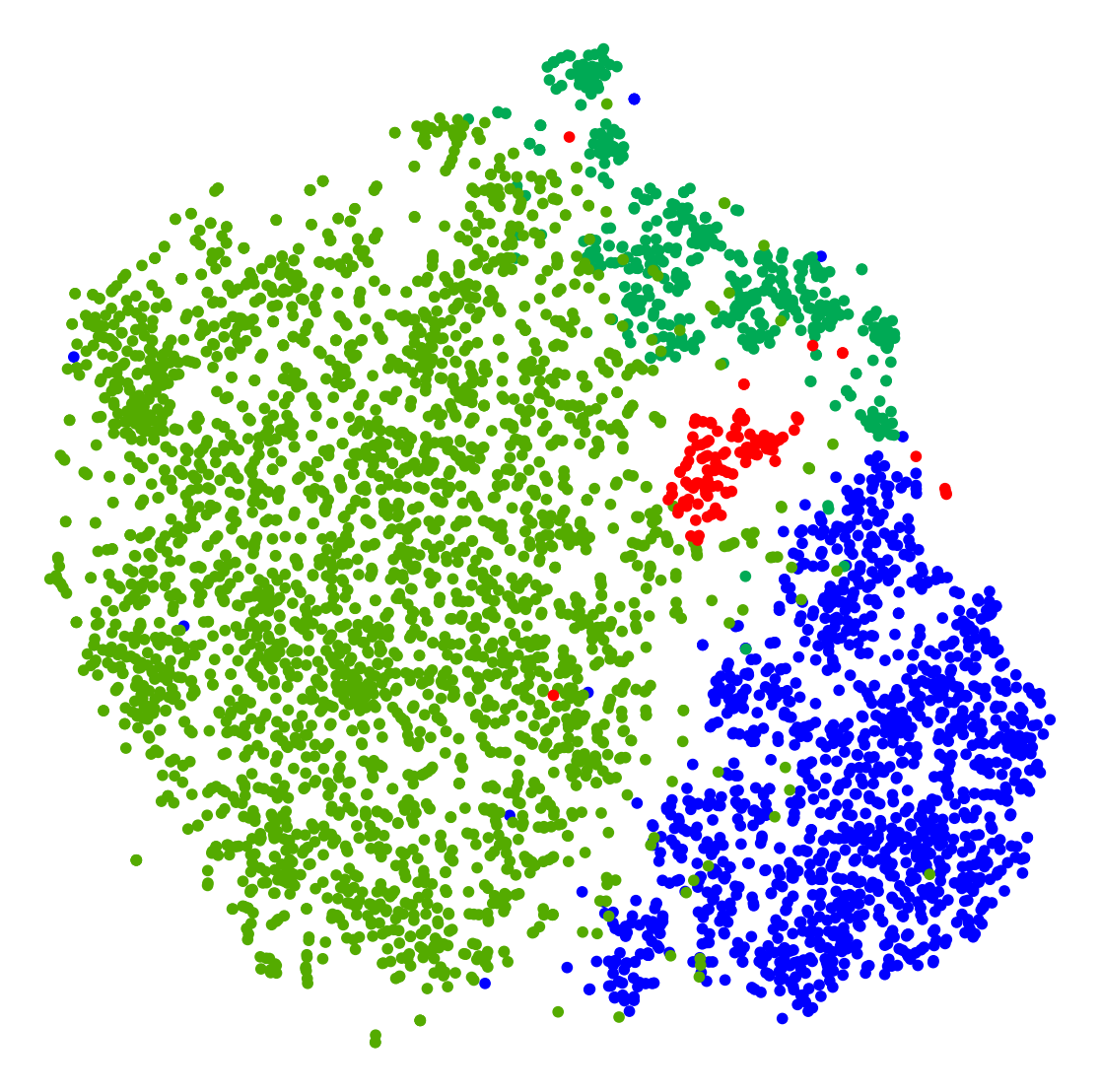}
  \caption{RCC}
  \label{fig:sub2}
\end{subfigure}%
\caption{COVID vs. Viral Pneumonia t-SNE visualization}
\label{fig:COVID-Viral_tsne}
\end{figure}

\begin{figure}[h!]
\centering
\begin{subfigure}[b]{.333\textwidth}
  \centering
  \includegraphics[width=\textwidth]{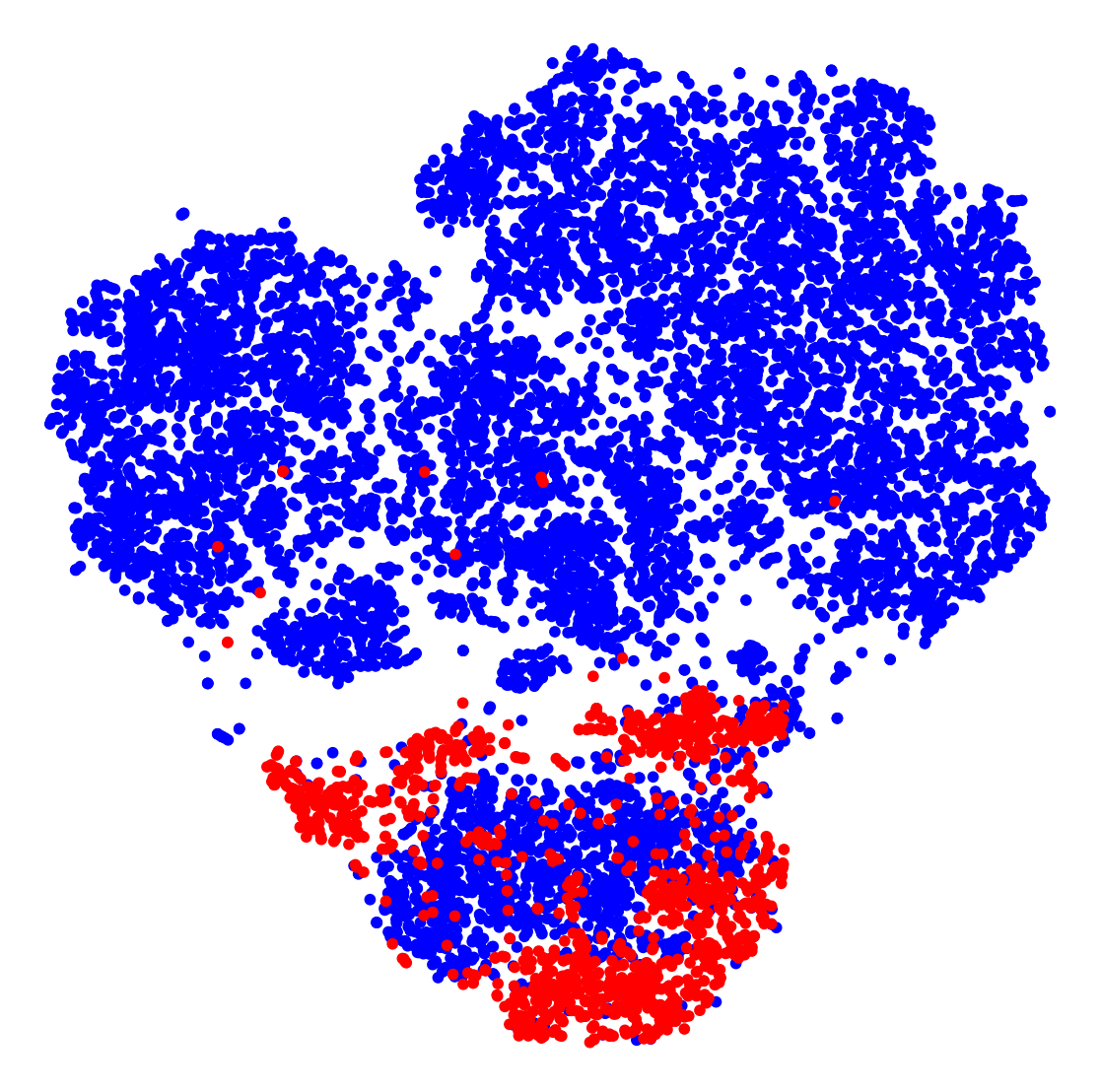}
  \caption{Ground truth}
  \label{fig:sub1}
\end{subfigure}%
\hfill
\begin{subfigure}[b]{.333\textwidth}
  \centering
  \includegraphics[width=\textwidth]{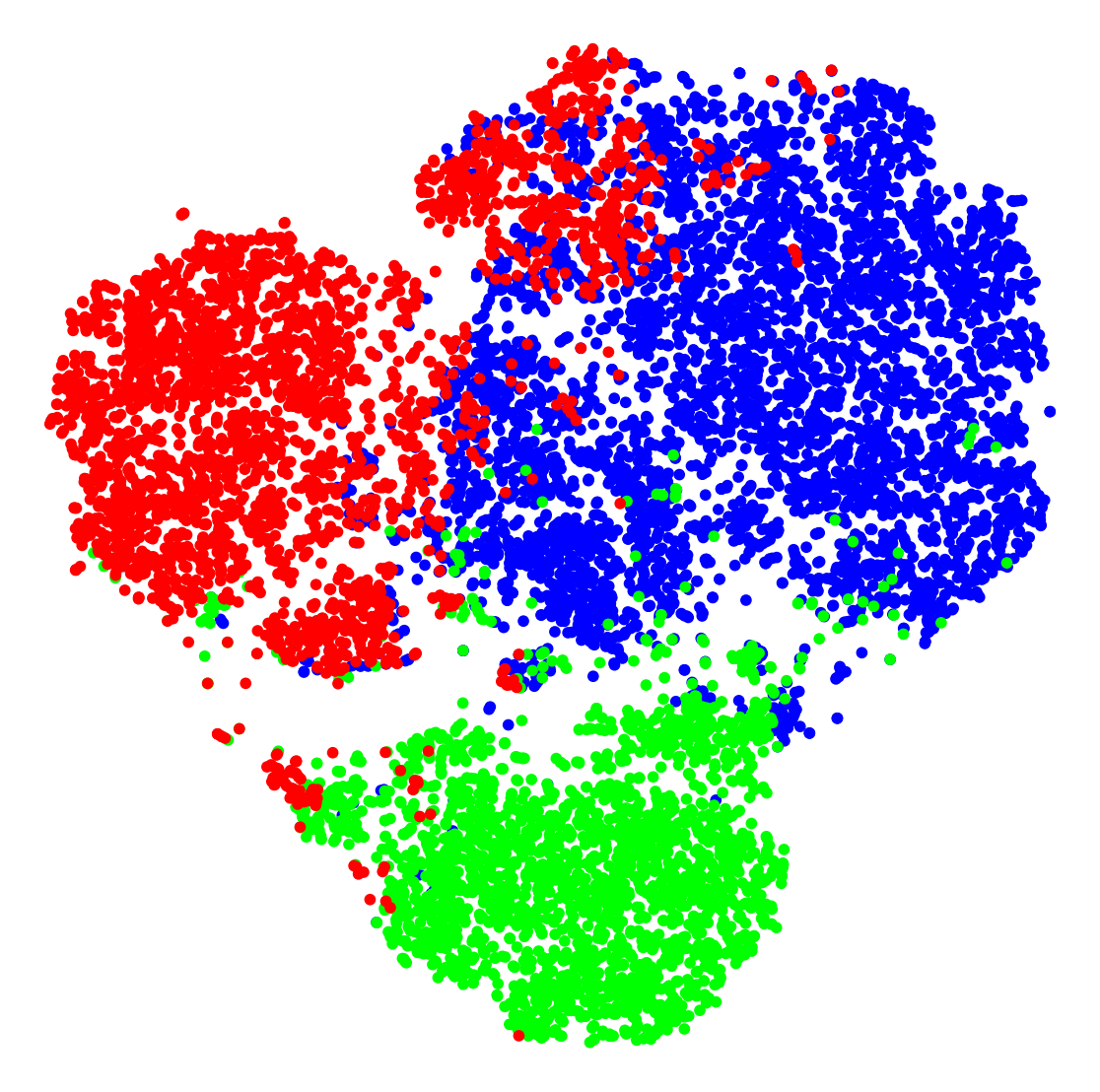}
  \caption{KM++}
  \label{fig:sub2}
\end{subfigure}%
\hfill
\begin{subfigure}[b]{.333\textwidth}
  \centering
  \includegraphics[width=\textwidth]{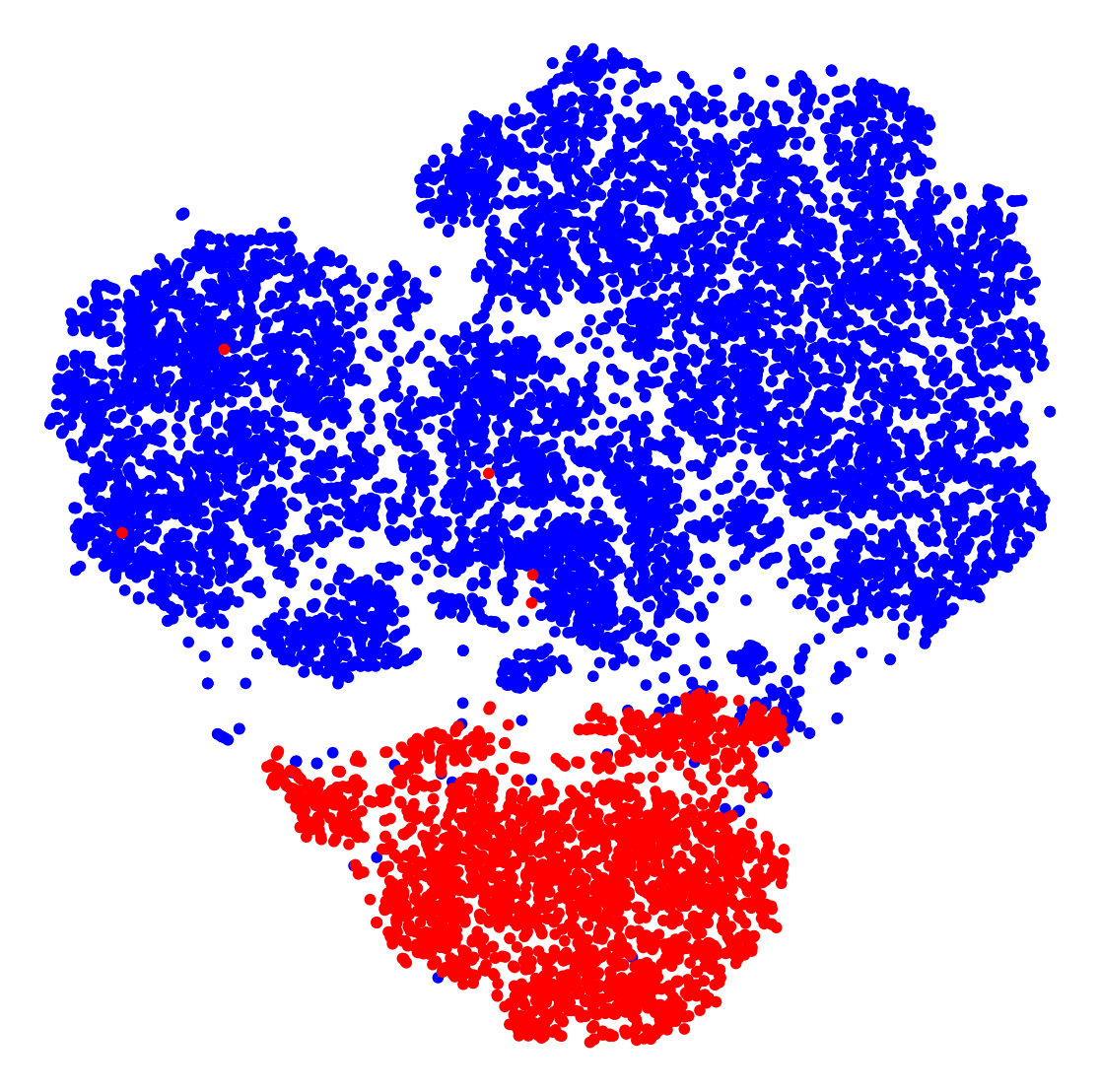}
  \caption{RCC}
  \label{fig:sub2}
\end{subfigure}%
\caption{Normal vs. Viral Pneumonia t-SNE visualization}
\label{fig:Normal-Viral_tsne}
\end{figure}

The main results are shown in Figure \ref{fig:ami_plots} and Tables \ref{table:1} and \ref{table:2}. The AMI score for KM++ experiments can be seen in figure \ref{fig:kmeans_solo} and the AMI score of RCC can be seen in figure \ref{fig:rcc_solo}. Both plots have identically scaled vertical axes: this allows us to compare the measured clustering performance of each experimental trial using the AMI metric. In Figure \ref{fig:kmeans_solo}, $k$ is the number of clusters used for KM++ and in Figure \ref{fig:rcc_solo}, $k$ the number of nearest neighbors for each data point in the m-KNN routine used during RCC. Tables \ref{table:1} and \ref{table:2} show the AMI values of the highest scoring trial run for each experiment. Clusterings and their AMI values were computed for four classification schemes: Three rounds of two-class comparisons were performed with the two element combinations of Normal, Viral Pneumonia, and COVID data and experiments were run with all three classes at once.

We have included t-SNE visualizations of the datasets with ground truth labelings and the best performing labelings for KM++ and RCC respectively. t-SNE allows us to glean some intuition into the shape and structure of the high dimensional dataset. While it does not give us a performance measure, it does allow us to visualize the results.

The results show that RCC consistently outperforms KM++ for three of the four experimental comparisons both in terms of peak performance and lack of sensitivity to the tested range of each algorithms' respective parameters. 

In the COVID-Normal comparison, RCC reaches a higher peak AMI than the KM++ experiment, but tapers off in its performance to a similar level as $k$ increases. We get some intuition as to why this may be occurring by looking at the t-SNE generated output in Figure \ref{fig:COVID-Normal_tsne(a)}. The low dimension representations of the ground truth labelings appear to be more heterogeneous than the other cases we consider, that is they do not separate nearly as well. This is likely a main reason why both KM++ and RCC exhibit poor clustering performance in this case. We can further notice that KM++ did not perform nearly as well as it does in the other cases when $k \in \{2, 3\}$, this is an indicator that the dataset is not trivially separable in its current form.




From the results, we observe that in general, RCC outperforms KM++ on the dataset. Nonetheless, we still run into the limitations of unsupervised clustering techniques. This is not a problem, rather it makes the experimentation quite interesting. RCC likely performs well because it uses a more refined measure of pairwise similarity than the $\ell_{2}$ distance. The connectivity structure constructed by m-KNN places a stricter requirement on determining if two data points are similar, thus the algorithm produces a better clustering of the data.

\section{Conclusions and Future Work}
\label{sec:conclusions}

The results from our experiments show that the RCC algorithm is able to correctly identify clusters of X-ray lung images in a promising fashion, and indeed a much more impressive manner than KM++. Despite RCC's general outperformance of KM++, both methods proved less effective in one of our experiments--COVID-Normal--indicating that while there is promise in the application of RCC in this context, further exploration is required. Because of this, our results should be viewed as a presentation of concept rather than a bona fide method for identifying illness. 

One direction for future work is to explore the effectiveness of different feature engineering techniques on this data set. Future works may also consider alternate dimensionality reduction techniques that may be more finely tuned to the COVID-19 dataset. Moreover, further exploration of the applicability of the RCC algorithm in this among other contexts is needed to fully understand the potential that this methodolgy presents. 

\section*{Acknowledgments}
We would foremost like to thank the PIC Math Program for this unique and rewarding opportunity. PIC Math is a program of the Mathematical Association of America (MAA). Support for this MAA program is provided by the National Science Foundation (NSF grant DMS-1722275) and the National Security Agency (NSA). We thank Dr. Shuai Zhang for proposing this interesting and important problem and for contributing useful suggestions throughout the project. We also thank Dr. Fred Park for his dedication, guidance, and support throughout this work. Lastly, we would also like to thank Qualcomm, Whittier College, NSF, MAA, and NSA.

\bibliographystyle{siamplain}
\bibliography{references}

\end{document}